\documentclass{article}
\usepackage[accepted]{icml2019}

\usepackage{xparse, xspace}
\usepackage{microtype}
\usepackage{graphicx}
\usepackage{subfigure}
\usepackage{booktabs} 
\usepackage{hyperref}
\usepackage{amsmath,amsfonts, amssymb, bm, amsthm, mathtools}

\DeclareMathOperator{\expec}{\mathbb{E}}
\DeclareMathOperator{\natu}{\mathbb{N}}
\DeclareMathOperator{\real}{\mathbb{R}}
\DeclareMathOperator{\posdef}{\mathbb{S}}
\DeclareMathOperator{\tensornorm}{\mathcal{TN}}
\DeclareMathOperator{\normal}{\mathcal{N}}

\DeclareMathOperator{\trace}{tr}
\DeclareMathOperator{\vecop}{vec}

\DeclarePairedDelimiterX{\infdivx}[2]{(}{)}{%
  #1\;\delimsize|\delimsize|\;#2%
}
\newcommand{\kld}[2]{\ensuremath{D_{KL}\infdivx{#1}{#2}}\xspace}

\begin{document}

\twocolumn[
\icmltitle{tvGP-VAE: Tensor-variate Gaussian Process Prior Variational Autoencoder}
\begin{icmlauthorlist}
\icmlauthor{Alex Campbell}{cam,ati}
\icmlauthor{Pietro Li\`{o}}{cam,ati}
\end{icmlauthorlist}
\icmlaffiliation{cam}{Department of Computer Science, University of Cambridge, Cambridge, United Kingdom}
\icmlaffiliation{ati}{The Alan Turing Institute, London, United Kingdom}
\icmlcorrespondingauthor{Alex Campbell}{ajrc4@cam.ac.uk}
\vskip 0.3in
]

\printAffiliationsAndNotice{}
\begin{abstract}
Variational autoencoders (VAEs) are a powerful class of deep generative latent variable model for unsupervised representation learning on high-dimensional data. To ensure computational tractability, VAEs are often implemented with a univariate standard Gaussian prior and a mean-field Gaussian variational posterior distribution. This results in a vector-valued latent variables that are agnostic to the original data structure which might be highly correlated across and within multiple dimensions. We propose a tensor-variate extension to the VAE framework, the tensor-variate Gaussian process prior variational autoencoder (tvGP-VAE), which replaces the standard univariate Gaussian prior and posterior distributions with tensor-variate Gaussian processes. The tvGP-VAE is able to explicitly model correlation structures via the use of kernel functions over the dimensions of tensor-valued latent variables. Using spatiotemporally correlated image time series as an example, we show that the choice of which correlation structures to explicitly represent in the latent space has a significant impact on model performance in terms of reconstruction.
\end{abstract}

\section{Introduction}
\label{introduction}
Multidimensional data generally refers to any dataset with more than two dimensions represented as a tensor. Tensors are a generalisation of matrices to multi-dimensional arrays with an arbitrary number of indices \cite{Kolda2009}. Real-world examples of tensor datasets range from meteorological measurements (\emph{latitude} $\times$ \emph{longitude} $\times$ \emph{altitude}), videos frames (\emph{row} $\times$ \emph{column} $\times$ \emph{channel} $\times$ \emph{time}), and medical imaging such as functional magnetic resonance imaging (fMRI) (\emph{height} $\times$ \emph{width} $\times$ \emph{depth} $\times$ \emph{time}). Commonly, tensor datasets are high-dimensional and interactions both within and across dimensions can be highly non-linear. Compact representations of tensor data are essential for exploratory analysis, density estimation, and feature learning for downstream discriminative tasks (e.g. classification, imputation). Discovering these representations is one of the main goals of unsupervised machine learning.

Deep generative latent variable models \cite{Oussidi2018}, particularly variational autoencoders (VAEs) \cite{Kingma2014}, are a powerful modelling framework for unsupervised non-linear dimensionality reduction. By specifying a prior distribution over unobserved latent variables, VAEs are able to learn features that represent high-dimensional data in a much simpler form. If latent variables are to be useful as features, then they must be able to model the full range of statistical properties of the input data \cite{Kwok2012a}. This includes how they vary across aspects of the data that \emph{a priori} have natural ordering and correlation such as space and time.

Oftentimes, however, VAEs are implemented with a univariate standard Gaussian prior and mean-field Gaussian variational posterior that assumes the latent variables are vector-valued and independent. As a result, the latent space is agnostic to the input data structure and unable to explicitly model correlations. Despite this, the hope is that a good fit to the data can still be achieved with a large enough number of latent variables coupled with powerful decoders \cite{Wang2019a}. The problem with this approach is that it can cause VAEs to ignore the task of learning useful representations and instead focus on estimating input data density: a phenomenon called posterior collapse \cite{Oord2018, Kim2018b}.

To allow for the explicit modelling of correlation structures in the latent space across an arbitrary number of dimensions, we propose the tensor-variate Gaussian process variational autoencoder (tvGP-VAE). The tvGP-VAE modifies the VAE framework by replacing the univariate Gaussian prior and Gaussian mean-field approximate posterior with tensor-variate Gaussian processes. As such, any input data tensor can be modeled as a sample from stochastic processes on tensors in a lower dimensional latent space. Furthermore, via the choice of kernel functions, any number of correlation structures in the input space can be encoded into the latent variables explicitly. We go on to show that by exploiting the properties of Kronecker separability, the tvGP-VAE is both computationally tractable and memory efficient and can be trained end-to-end using amortised stochastic variational inference.

Our aim in developing the tvGP-VAE framework is to explore whether tensor-valued latent variables are better suited for learning useful representation of multidimensional data than their vector-valued counter parts. Although recent work has made progress generalising convolutions to work efficiently on high-dimensional tensor data \cite{Kossaifi2020} the use of tensor-variate probability distributions for deep learning remains a relatively underdeveloped area.

The rest of this paper is organized as follows. Section \ref{background} introduces the tensor-variate Gaussian distribution, tensor-variate Gaussian processes and VAEs. Section \ref{tensor variate gaussian process prior variational autoencoder} formally develops the tvGP-VAE model outlining the modifications that need to be made to the generative process and inference procedure of the standard VAE. Section \ref{related work} discusses related work. Finally, experimental results and conclusions are set-out in Section \ref{experiments} and Section \ref{conclusion} respectively.

\section{Background}
\label{background}

\subsection{Tensors}
\label{tensors}
A tensor is multidimensional extension of an array. The order of a tenor is the number of dimensions (also referred to as its way). Each dimension is called a mode. Let $\bm{\mathcal{X}} \in \real^{D_1 \times \dots \times D_M}$ denote an order-$M$ tensor where $D_m \in \natu$ is the number of dimensions in the $m$-th mode and $D=\prod_{m=1}^M D_m$ the total dimensionality. Furthermore, let $\mathcal{X}_{i_1, \dots, i_M}$ denote the $(i_1,\dots,i_M)$-th element of $\bm{\mathcal{X}}$. For a more detailed overview of tensors see \citet{Kolda2009}.

Throughout this paper zeroth-order tensors or scalars are denoted in lowercase, e.g., $x$; first-order tenors or vectors in bold lowercase, e.g., $\bm{x}$; second-order tensors or matrices in bold uppercase, e.g.,  $\bm{X}$; and third-order tensors and above boldface Euler script, e.g., $\bm{\mathcal{X}}$.

\subsection{Tensor-variate Gaussian distribution}
\label{tensor gaussian}
A tensor-variate Gaussian distribution \cite{Ohlson2013, Xu2019} is a probability distribution over random tensor $\bm{\mathcal{X}} \in \real^{D_1 \times \dots \times D_M}$ of order-$M$ with probability density defined
\begin{align}
\begin{split}
p(\bm{\mathcal{X}}) &= \tensornorm(\bm{\mathcal{X}}; \bm{\mathcal{M}}, \bm{\Sigma}^{(1)}, \dots, \bm{\Sigma}^{(M)}) \\
&= \frac{\exp \left[ -\frac{1}{2} \bm{\mathcal{Q}}\mathop{\times}\limits_{m=1}^{M} \left( \mathop{\circ}\limits_{m=1}^{M}\bm{\Sigma}^{(m)-1}\right)\mathop{\times}\limits_{m=1}^{M} \bm{\mathcal{Q}} \right]}{(2 \pi)^{\frac{D}{2}}\prod_{m=1}^M|\bm{\Sigma}^{(m)}|^{\frac{D}{2D_m}}}
\end{split}
 \label{def tensor gaussian}
\end{align}
such that $\bm{\mathcal{Q}}=(\bm{\mathcal{X}}-\bm{\mathcal{M}})$ where $\bm{\mathcal{M}} \in \real^{D_1 \times \dots \times D_M}$ is the mean and $\bm{\Sigma}^{(m)} \in \posdef^{D_m}_{++}$ is the mode-$m$ covariance. The operator $|.|$ denotes the matrix determinant, $\times_m$ the mode-$m$ product between a tensor and matrix, and $\mathop{\circ}$ is the outer product of matrices.

The matrix-variate Gaussian is equivalent to a multivariate Gaussian distribution such that 
\begin{align}
p(\vecop(\bm{\mathcal{X}})) = \normal(\vecop(\bm{\mathcal{M}}), \mathop{\otimes}\limits_{m=3}^{1})
\label{vec property}
\end{align}
where $\vecop(.)$ is the vectorisation operator and $\otimes$ is the Kronecker product. When $M=1$ the tensor-variate Gaussian reduces to the multivariate Gaussian and when $M=2$ it reduces to the matrix-variate Gaussian \cite{Gupta2000, Dawid1981}.

\subsection{Tensor-variate Gaussian process}
\label{tensor gaussian process}
Gaussian processes \cite{Rasmussen2004} are a class of probabilistic models specifying a distribution over function spaces.
 Given an index set $\mathbb{I} \coloneqq \mathbb{I}_1 \times \dots \times \mathbb{I}_M$, the stochastic process $\{\mathcal{X}(\bm{i}): \bm{i} \in \mathbb{I} \}$ is a tensor-variate Gaussian process \cite{Chu2009, Xu, Zhao} if for every finite set of indices $\bm{I} \coloneqq \bm{I}_1 \times \dots \times \bm{I}_M$, where $\bm{I}_m \subset \mathbb{I}_m$,  the tensor with elements defined $\bm{\mathcal{X}} \coloneqq \{\mathcal{X}(\bm{i}_{n}) : \bm{i}_n \in \bm{I} \}$ is a order-$M$ tensor Gaussian random variable
\begin{align}
\begin{gathered}
\bm{\mathcal{X}} \sim \mathcal{TGP}(m(.), \kappa(., .)) \\
 \iff \\
 p(\bm{\mathcal{X}}) = \tensornorm(\bm{\mathcal{X}}; \bm{\mathcal{M}}, \bm{\Sigma}^{(1)}, \dots, \bm{\Sigma}^{(M)})  
\end{gathered}
\label{def tensor gaussian process}
\end{align}
where $\bm{\mathcal{M}}\coloneqq\{m(\bm{i}_n): \bm{i}_n \in \bm{I}\}$ is the mean function and $\bm{\Sigma}^{(m)}\coloneqq\{\kappa^{(m)}(i, i'; \lambda^{(m)}): i, i' \in \bm{I}_m\}$ is the covariance function with kernel $\kappa(., .)$ with parameters $\lambda^{(m)}$.

\subsection{Variational autoencoder}
\label{variational autoencoder}
Variational autoencoders (VAEs) (Rezende 2014; Kingma 2014) are a framework for efficiently learning probabilistic latent variable models leveraging the representative power of deep neural networks (DNN) for the generative model and inference. 

\paragraph{Generative model} Let $\bm{\mathcal{X}} \in \real^{D_1 \times \dots, \times D_M}$ denote high-dimensional data and $\bm{\mathcal{Z}} \in \real^{D_1' \times \dots, \times D'_{M'}}$ its corresponding low-dimensional latent representation, $D' \ll D$.

A standard VAE models the joint distribution $p_\theta(\bm{\mathcal{X}}, \bm{\mathcal{Z}})$ via the generative process
\begin{align}
\bm{\mathcal{Z}} &\sim p_\theta(\bm{\mathcal{Z}}) \label{prior}\\ 
\bm{\mathcal{X}} &\sim p_\theta(\bm{\mathcal{X}}|\bm{\mathcal{Z}}) = p(\bm{\mathcal{X}}| g_\theta(\bm{\mathcal{Z}})) 
\label{conditional likelihood}
\end{align}
where $g_\theta(.)$ is a DNN called the decoder (or generative network) that parameterises a tractably computable likelihood for $\bm{\mathcal{X}}$. 

\paragraph{Inference} A model for the data likelihood $p_\theta(\bm{\mathcal{X}})$ is learnt using amortised variational inference to approximate the intractable posterior with a freely chosen tractable density $q_\phi(\bm{\mathcal{Z}}|\bm{\mathcal{X}}) \approx p(\bm{\mathcal{Z}}|\bm{\mathcal{X}})$ that is parameterised by a DNN $f_\phi(.)$ called the encoder (or inference network)
\begin{align}
q_\phi(\bm{\mathcal{Z}}|\bm{\mathcal{X}})=q(\bm{\mathcal{Z}}|f_\phi(\bm{\mathcal{X}})). 
\label{posterior}
\end{align}
Inference is equivalent to maximising a lower bound on the data log-likelihood, the evidence lower bound (ELBO) \cite{Saul1996}, with respect to the parameters $\phi$ and $\theta$
\begin{multline}
\mathcal{L}_{\text{ELBO}} \coloneqq \underbrace{\expec_{q(\bm{\mathcal{Z}}|\bm{\mathcal{X}})}\left[ \log p_\theta(\bm{\mathcal{X}}|\bm{\mathcal{Z}})\right]}_\text{$\mathcal{L}_R$} \\ -\underbrace{\kld {q_\phi(\bm{\mathcal{Z}}|\bm{\mathcal{X}})}{p_\theta(\bm{\mathcal{X}})}}_\text{$\mathcal{L}_C$} \leq \log p_\theta(\bm{\mathcal{X}})
\label{elbo}
\end{multline}
 where $\kld{.}{.}$ represents to the Kullback-Leibler divergence, $\mathcal{L}_R$ the reconstruction loss and $\mathcal{L}_C$ is the complexity loss.

\paragraph{Independent vector-valued latent space} 
VAEs are usually implemented assuming the latent variables are first-order tenors (or vectors), $\bm{Z}$, with dimensionality $D'_1 = K$ sampled from a standard Gaussian prior. Combined with a mean-field Gaussian for the posterior, both distributions factor fully across dimensions:
\begin{align}
p_{\theta}(\bm{Z}) &= \prod_{k=1}^{K} \normal(z_{k}; 0, 1) \label{fully factored prior}\\ 
q_{\phi}(\bm{Z}|\bm{\mathcal{X}}) &= \prod_{i=1}^{K} \normal(z_{k}; f_{\phi}(\bm{\mathcal{X}})) \label{guassian mean field}\\
f_{\phi}(\bm{\mathcal{X}}) &= \{\mu_{i}(\bm{\mathcal{X}}), \sigma_{i}(\bm{\mathcal{X}}) \}_{k=1}^{K}. \notag
\end{align}
Although this ensure that $\mathcal{L}_C$ can be calculated analytically and $\mathcal{L}_R$ can be approximated by Monte Carlo integration of a reparameterised approximate variational posterior, the latent variables $\bm{Z}$ are independent and have no explicit correlation structure.

\section{Tensor-variate Gaussian process prior variational autoencoder (tvGP-VAE)}
\label{tensor variate gaussian process prior variational autoencoder}

In this section we build upon the standard VAE set out in Section \ref{variational autoencoder} and propose the tvGP-VAE. The tvGP-VAE framework allows for the learning of tensor-valued latent variables that can explicitly model correlation structures from the input data in a lower dimensional space. We focus here on image time series but it is trivial to extend the model to higher order tensor data with any correlation structure that can be specified with a kernel.

\subsection{Problem statement}
Let the fourth-order tensor $\bm{\mathcal{X}} \in \real^{C \times W \times H \times T}$ denote an image time series with spatial dimensions $W \times H$, temporal dimension $T$, and colour channels $C$. Furthermore let the forth-order tensor $\bm{\mathcal{Z}} \in \real^{K \times W' \times H' \times T'}$ denote a low-dimensional latent representation of $\bm{\mathcal{X}}$. As in the standard VAE framework, we wish to learn $K$ representations of $\bm{\mathcal{X}}$ to be used in a downstream task (e.g. classification, denoising, imputation, disentanglement). However in contrast to Eq. \eqref{fully factored prior} and Eq. \eqref{guassian mean field}, we assume that the prior and posterior only factor over the first mode
\begin{align}
p_{\theta}(\bm{\mathcal{Z}}) &= \prod_{k=1}^{K}p_{\theta}(\bm{\mathcal{Z}}_k) \label{new prior}\\ 
q_{\phi}(\bm{\mathcal{Z}}|\bm{\mathcal{X}}) &= \prod_{k=1}^{K}  q_{\phi}(\bm{\mathcal{Z}}_k|\bm{\mathcal{X}}) \label{new posterior}
\end{align}
where $\bm{\mathcal{Z}}_k \in \real^{W' \times H' \times T'}$. This gives us the flexibility to choose which modes, and their respective dimensionality, to represent from $\bm{\mathcal{X}}$. (Following the notation from Section \ref{tensors}, this is equivalent to setting $M=4$, $D_1=C$, $D_2=W$, $D_3=H$, $D_4=T$ for $\bm{\mathcal{X}}$ and $M'=4$, $D_1'=K$ $D_2'=W'$, $D_3'=H'$, $D_4'=T'$ for $\bm{\mathcal{Z}}$ respectively where $D' \ll D$.)

\subsection{Tensor-variate Gaussian process prior}
To allow for the specification of correlation structures over each mode of $\bm{\mathcal{Z}}_k$, we place a tensor-variate Gaussian process prior over the index set
\begin{align}
\bm{I} &\coloneqq \underbrace{\{1, \dots, W'\}}_\text{$\bm{I}_1$} \times \underbrace{\{1, \dots, H'\}}_\text{$\bm{I}_2$} \times \underbrace{\{1, \dots, T'\}}_\text{$\bm{I}_3$} \notag
\end{align}
such that Eq. \eqref{new prior} becomes
\begin{align}
p_\theta(\bm{\mathcal{Z}}_k) &=\tensornorm (\bm{\mathcal{Z}}_k; \bm{\mathcal{O}}_k, \bm{\Omega}^{(1)}_k, \bm{\Omega}^{(2)}_k, \bm{\Omega}^{(3)}_k) \label{tv prior}
\end{align}
where $\bm{\mathcal{O}}_k \coloneqq \{m_k(\bm{i})=0 : \forall \bm{i} \in \text{\bf{I}} \}$ is a constant zero tensor and $\bm{\Omega}^{(m)}_k \coloneqq \{\kappa^{(m)}_k(i, i'; \lambda^{(m)}_k): i, i' \in \bm{I}_m \}$ is the order-$m$ covariance matrix for $k=1, \dots K$. 

\paragraph{Kernel functions}
The choice of kernels are primarily based on \emph{a priori} knowledge about correlation structures in the input space $\bm{\mathcal{X}}$. In our case, to model spatial and temporal correlation we choose the squared exponential (or exponentiated quadratic) kernel for each mode of $\bm{\mathcal{X}}$
\begin{align}
\begin{gathered}
\kappa^{(m)}_k(i, i'; \lambda^{(m)}) = (\sigma_k^{(m)})^2 \exp \left( -\frac{(i-i')^2}{2l{_k}^{(m)}} \right) 
\end{gathered}
\end{align}
where $\lambda_k^{(m)}= \{\sigma_k^{(m)}, l_k^{(m)}\}$ for $k=1, \dots, K$. The signal standard deviation $\sigma_k^{(m)}>0$ controls the variability from the mean function and the length scale $l_k^{(m)}>0$ controls variability over the domain $\bm{I}_m $. 

\subsection{Tensor-variate Gaussian process posterior} 
We choose a tensor-variate Gaussian process for the approximate posterior in Eq \eqref{new posterior}. The encoder $f_{\phi}(.)$ network learns the parameters in two stages. First, an intermediate representation $\bm{\mathcal{H}}_k$ indexed by $\bm{I}$ is learnt for every $k=1, \dots, K$. Second, mean and covariance parameters are output for each index $i$ of every mode $\bm{I}_m$ of $\bm{\mathcal{H}}_k$ (the total number of parameters per index is discussed below). This defines a tensor-variate Gaussian process over $\bm{I}$ such that
\begin{align}
 q_\phi(\bm{\mathcal{Z}}_k|\bm{X}) &=\tensornorm(\bm{\mathcal{Z}}_k;  f_\phi(\bm{\mathcal{X}}))  \notag \\
 f_\phi(\bm{\mathcal{X}}) &= \{ \bm{\mathcal{M}}_k(\bm{\mathcal{X}}),\bm{\Sigma}_k^{(1)}(\bm{\mathcal{X}}),  \\ &\qquad \bm{\Sigma}_k^{(2)}(\bm{\mathcal{X}}), \bm{\Sigma}_k^{(3)}(\bm{\mathcal{X}})\}_{k=1}^K. \notag
\end{align}
We henceforth drop the dependency of the distributional parameters on $\bm{\mathcal{X}}$ for notational clarity.

\paragraph{Posterior parameter constraints} The flexibility of using a tensor-variate Gaussian for the posterior $ q_\phi(\bm{\mathcal{Z}}_k|\bm{X})$ comes at the cost of increased complexity as the inference network $f_\phi(.)$ must now output $K(W'H'T' + W'(W'+1)/2 + H'(H'+1)/2 + T'(T'+1)/2)$ local variational parameters. The cost of estimating the mean tensor $\bm{\mathcal{M}}_k$ being dominant for each $k=1, \dots, K$. Furthermore, evaluating the probability density requires inverting potentially dense covariance matrices which is cubic in time $\mathcal{O}(K((W')^3 + (H')^3+ (T')^3))$. To reduce the number of parameters and sampling time complexity, we place a low-rank constraint on the mean and structured sparsity constraints on each covariance matrix. 

\paragraph{Low-rank mean tensor} The mean tensor $\bm{\mathcal{M}}_k$ is given a low rank structure by constraining it to be Kronecker separable \cite{Dees2019a, Allen2010, Wang2019a}. As such, the the mean can be be decomposed into the outer product of first-order tensors (or column vectors)
\begin{align}
\bm{\mathcal{M}}_k = \bm{m}_k^{(1)}\mathop{\circ}\bm{m}_k^{(2)}\mathop{\circ}\bm{m}_k^{(3)}
\label{low rank mean}
\end{align}
where $\bm{m}_k^{(m)} \in \real^{\bm{I}_m}$ is the order-$m$ mean. This reduces the numbers of local variational parameters for the mean tensor to $K(W' + H' + T')$.

\paragraph{Sparse covariance matrices}
To avoid having to parameterise and invert potentially dense covariance matrices for each mode of $\bm{\mathcal{Z}}_k$, we instead directly parameterise the Cholesky factors of precision matrices $\bm{\Sigma}_k^{(m)-1} \coloneqq \bm{L}_k^{(m)} (\bm{L}_k^{(m)})^\top$ where $\bm{L}_k^{(m)}$ is a lower triangle matrix. To induce sparsity, the correlation structures for each mode of $\bm{\mathcal{Z}}_k$ are constrained to follow a first-order autoregressive process. This is motivated by the fact that the order of $\bm{\mathcal{Z}}$ models either spatial or temporal correlation in $\bm{\mathcal{X}}$.  As such, the Cholesky factor $\bm{L}_k^{(m)}$ is parameterised
\begin{align}
\bm{L}_k^{(m)} &= \{ l^{(m, k)}_{i, i}, l^{(m, k)}_{i+1, i} \}_{i = 1}^{D_m'}
\label{low rank cov}
\end{align}
where $l^{(m, k)}_{i, j}$ is the $ij$-th local variational parameter for $\bm{L}_k^{(m)}$ for $k=1, \dots, K$. By construction, each $\bm{L}_k^{(m)}$ is bidiagonal resulting in a precision matrix $\bm{\Sigma}_k^{(m)-1}$ which is banded and sparse and thus capable of being evaluated in linear time. This reduces the number of local variational parameters for each covariance matrix to $K(2D_m'-1)$.

\paragraph{Output of inference network} Given the constraints on the mean and covariance matrices from Eq. \eqref{low rank mean} and Eq. \eqref{low rank cov} respectively, the inference network outputs $3$ parameters
\begin{align}
f_\phi(\bm{\mathcal{H}}_k) = \{m_i^{(m, k)}, l^{(m, k)}_{i, i}, l^{(m, k)}_{i+1, i}\}
\end{align}
for each index of $i=1, \dots, \bm{I}_m$ of every mode $m=1, 2, 3$ where $m_i^{(m, k)} \in \bm{m}^{(m)}_k$ and $ l^{(m, k)}_{i, i} \in \bm{L}_k^{(m)}$.

\begin{figure}[H]
\begin{center}
\centerline{\includegraphics[width=\columnwidth]{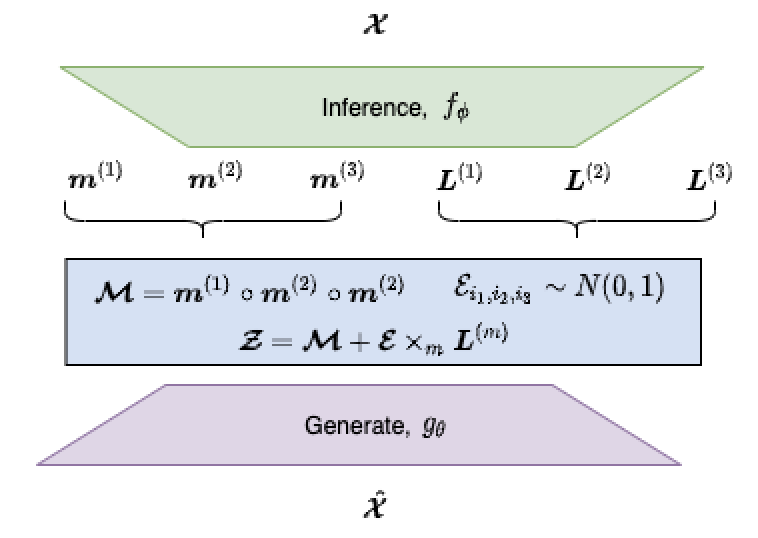}}
\caption{tvGP-VAE inference and generative model. The tvGP-VAE allows for tensor-valued latent variables capable of explicitly modelling correlation structures.}
\label{model}
\end{center}
\end{figure}

\subsection{Evidence lower bound} 
\paragraph{Complexity loss} Using Eq. \eqref{vec property} and the KL-divergence between two multivariate Gaussian's, the complexity loss $\mathcal{L}_C$ retains a closed form solution
\begin{align}
\begin{aligned}
\mathcal{L}_C &=\frac{1}{2} \sum_{k=1}^K \Big(\prod_{m=1}^{3}\trace[\bm{\Omega}_k^{(m)-1}\bm{\Sigma}_k^{(m)}] \\ 
& + \vecop(\bm{\mathcal{M}}_k)^\top \mathop{\otimes}\limits_{m=3}^{1}\bm{\Omega}_k^{(m)-1}\vecop(\bm{\mathcal{M}}_k) \\
& + \sum_{m=1}^3 (\log |\bm{\Omega}_k^{(m)}| - \log |\bm{\Sigma}_k^{(m)}|) -D'\Big).
\end{aligned}
\end{align}
See Supplementary Materials for a full derivation.

\paragraph{Posterior sampling}
Reparameterised samples can be drawn from the posterior using auxiliary tensor noise $\bm{\mathcal{E}}$
\begin{align}
\bm{\mathcal{Z}}_k &= \bm{\mathcal{M}}_k + \bm{\mathcal{E}} \mathop{\times}\limits_{m=1}^{3} \bm{L}_k^{(m)} 
\end{align}
where $\bm{\mathcal{E}} \coloneqq \{\epsilon: \epsilon \sim \normal(0, 1) \}$ is just $D'$ samples from a standard Gaussian reshaped into a tensor, and $\times_m$ is the mode-$m$ product. Gradients can therefore flow through $\mathcal{L}_{\text{ELBO}}$ allowing for be optimisation using stochastic variational inference. See Figure \ref{model} for an overview of a model overview.

\section{Related work}
\label{related work}

\paragraph{Structured VAE} Our model falls within the branch of ``structured VAEs" that aim for richer latent representations by relaxing the assumption of a fully factored Gaussian distribution for use in either the prior or the approximate variational posterior, or both. Previous attempts at making the approximate variational posterior richer include inverse autoregressive flows \cite{Kingma2017a}, Gaussian mixtures \cite{Nalisnick2016}. For richer priors previous proposals include non-parametric stick-breaking processes \cite{Nalisnick2016},  von Mises-Fisher distribution \cite{Davidson2018}, and autoregressive priors \cite{Razavi2019}.

\paragraph{Matrix-variate VAE} In the context of VAEs, only matrix latent representations have been considered to-date. Both \citet{Wang2019a} and \citet{Li2020} use matrix-variate Gaussian distributions for the approximate variational posterior distribution in order to model the spatial correlation structures between the rows and columns of image pixels. In contrast to our approach, however, both works assume a standard Gaussian prior limiting the type of correlation structures that can be represented in the latent space.

\paragraph{Gaussian process VAE} \citet{Casale2018} was the first to propose using a Gaussian process prior within the VAE framework to model correlation structures in the latent space. Closest to our approach, however, is the Gaussian process VAE of \citet{Fortuin2019}. Although the focus is on data imputation for multivariate time series, they use a Gaussian process prior over the latent space of a VAE to model temporal correlation in image time series as well as parameter constraints to induce sparsity in variational parameters. The latent representations, however, are limited to vectors allowing for no explicit spatial structure.

\section{Experiments}  
\label{experiments}
Experiments were performed on the image time series dataset Sprites \cite{li2018disentangled}. We contrasted 3 variations of the tvGP-VAE model, each explicitly modelling a different combination of modes of the input data, with a standard VAE in terms of quantitative and qualitative performance. Quantitative assessment is in terms of the reconstruction loss/ negative log likelihood. Qualitative assessment is accomplished through visualisation of data reconstructions. The aim was to asses how explicitly modelling different correlation structures, via different orders of tensor latent variables, effect reconstruction.


\subsection{Sprites dataset} 
The Sprites dataset consists of 10,000 sequences of animated characters with varying clothes, hair and skin colour, each performing a different action. The spatial dimensions are $W \times H = 64 \times 64$, the temporal dimension is $T=8$, and the number of channels are $C=3$. All images were normalised to $[0, 1]$.
 
\subsection{Implementation}
 
\paragraph{Models}
The flexibility of the tvGP-VAE framework was used to explore how different orders of tensor-valued latent variables compared to their vectored-valued counterparts in terms of reconstruction quality. We fixed the number of representations to $K=4$ and modelled explicit spatio-temporal correlation with a third-order tensor $\bm{\mathcal{Z}}_k \in \real^{W' \times H' \times T'}$, spatial correlation with a second-order tensor $\bm{\mathcal{Z}}_k \in \real^{W' \times H'}$, temporal correlation with a first-order tensor $\bm{\mathcal{Z}}_k \in \real^{T'}$, and no explicit correlation structure with a zero-order tensor $\bm{\mathcal{Z}}_k \in \real$, which corresponds to the standard VAE. The number of dimensions in each mode was fixed to $W'=H'=T'=4$. For the kernel functions $\kappa^{(m)}_k(.)$ in the prior, we use the same length scale and standard deviation for all $m$ modes across all $k$ dimensions such that $\sigma_k^{(m)}=l_k^{(m)}=1$.
 
 \paragraph{Neural network architecture} The encoder $f_{\phi}(.)$ and the the decoder $g_\theta(.)$ networks were implemented using convolutional/transpose convolutional layers to downsample/upsample the input data. We use 2D spatial convolutions to extract spatial features for each time point followed by 1D temporal convolutions to integrate temporal information into the features for each spatial location. All convolutions used a kernel size of 4 and were followed by batch normalisation and a rectified linear activation. To make the comparison as fair as possible, we kept the architectures the same across all models except for the output/input layers of the encoder/decoder which used different sized fully connected layers to account for the difference in local variational parameters/latent dimensions. See the Supplementary Materials for further details on neural network architectures.

\paragraph{Training and optimisation}
The data was randomly split into training/validation/test sets ($0.8/0.1/0.1$) and trained with a batch size of $50$ for $500$ epochs, which was sufficient for convergence of the loss, $\mathcal{L}_{\text{ELBO}}$. Every 10 epochs the loss was evaluated on the validation set and training stopped if it failed to decrease $5$ consecutive times in a row. We performed optimisation of the model parameters $\phi$ and $\theta$ using the Adam algorithm \cite{kingma2014adam} with a learning rate of $1e-3$.

\subsection{Results}
Table \ref{quant results} summarises the reconstruction loss of all models on the Sprites test dataset. The tvGP-VAE with explicit temporal structure as well as the model with explicit spatial structure both outperformed the standard VAE achieving lower reconstruction losses. The worst performing model was the spatio-temporal tvGP-VAE. 

\begin{table}[H]
\caption{Negative log-likelihood (reconstruction loss) of VAE and tvGP-VAE on Sprites test dataset. Reported values are means and their respective standard errors. The best scores are shown in bold (lower is better).}
\label{quant results}
\begin{center}
\begin{small}
\begin{tabular}{lrrrrl}
\toprule
Model & \multicolumn{4}{c}{Latent dimensions} & Neg. log likelihood \\
\cmidrule(r){2-5} 
& $K$ & $W'$ & $H'$ & $T'$ & \\
\midrule
VAE & 4 & \_ & \_ & \_ & $90, 616 \pm 3.8$ \\
    & 4 & 4 & 4 & 4 & $90, 684 \pm 15.7$ \\
tvGP-VAE & 4 & 4 & 4 & \_ & $90, 587 \pm 11.3$ \\
 & 4 & \_ & \_ & 4 &  $\bm{90, 578 \pm 6.2}$ \\
\bottomrule
\end{tabular}
\end{small}
\end{center}
\end{table}

\begin{figure}[ht]
\begin{center}
\centerline{\includegraphics[width=\columnwidth]{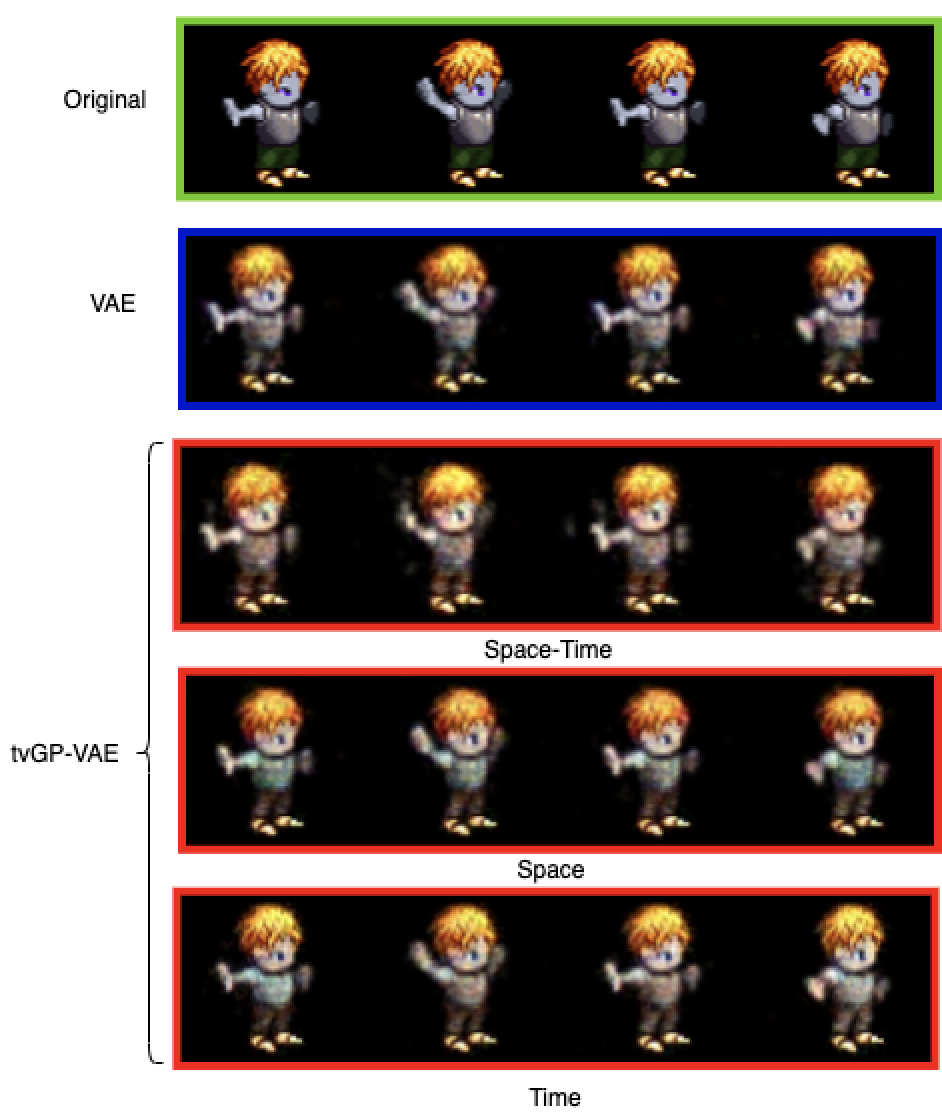}}
\caption{Reconstructions of a single observation from the Sprites test dataset (first four time points). The first row (green) is the original image, the second row (blue) is a reconstruction from the standard VAE wheres the last three rows (red) are from the tvGP-VAE model with spatio-temporal, spatial, and temporal structure in the latent space respectively.}
\label{qual results}
\end{center}
\end{figure}

From Figure \ref{qual results}, the spatio-temporal tvGP-VAE reconstruction clearly has less detail and higher uncertainty around the areas of motion despite having the ability to model pixel correlation over time. On the other hand, the spatial version of the tvGP-VAE captures the most detail out of all the models which intuitively makes sense as it can explicitly account for row and column pixel correlations. Despite the greater level of detail, the temporal tvGP-VAE outperforms its spatial counterpart as the cost of incorrectly reconstructing an area of motion is more costly.   

Although these results suggest that tensor-valued latent variables, capable of modelling explicit correlation structures in the input, can help improve reconstruction, the increased flexibility comes at a cost. The tensor-variate Gaussian process prior increases the variance within each mode via the kernel length scale parameter $l_k^{(m)}$. In the case of the spatio-temporal tvGP-VAE, this adds three extra hyperparameters per dimension $k$ that require careful tuning. 

\section{Conclusion}
\label{conclusion}
We have introduced a tensor-variate Gaussian process variational autoencoder (tvGP-VAE) to allow for the explicit modelling of correlation structures in a tensor-valued latent space. The oversimplifying assumptions of a standard Gaussian prior and mean-field Gaussian approximate posterior from the VAE framework are replaced with tensor variate Gaussian processes allowing for the specification of arbitrary correlation structures in the latent space via kernel functions. We demonstrated using an image time series dataset that the order of tensor used as the latent representation has an effect on reconstruction quality. Future work will include experiments on more tensor datasets with different correlation structures as well as investigating the effect of changing the dimensionality of each mode in the latent space on reconstruction loss. Furthermore, different choices of kernel functions and tuning of the length scale parameters will be considered.  

\section*{Acknowledgements}
This work was supported by The Alan Turing Institute under the EPSRC grant EP/N510129/1.

\clearpage

\bibliography{bibliography}
\bibliographystyle{icml2019}



\end{document}